\documentclass[10pt,twocolumn,letterpaper]{article}

\usepackage{cvpr}
\usepackage{times}
\usepackage{epsfig}
\usepackage{graphicx}
\usepackage{amsmath}
\usepackage{amssymb}
\usepackage{multirow}
\usepackage{bbm}
\usepackage{rotating}
\usepackage{array}
\usepackage{threeparttable}
\allowdisplaybreaks[4]

\usepackage{epstopdf}
\usepackage{subfigure}
\usepackage{amsmath}
\usepackage{bm}
\usepackage{url}
\usepackage{amsfonts,amssymb}
\usepackage{algorithm}
\usepackage{algorithmic}

\usepackage[breaklinks=true,bookmarks=false]{hyperref}

\cvprfinalcopy % *** Uncomment this line for the final submission

\def\cvprPaperID{****} % *** Enter the CVPR Paper ID here

\begin{document}

%%%%%%%%% TITLE
\title{Mitigate Bias in Face Recognition using Skewness-Aware Reinforcement Learning}

\author{Mei Wang\textsuperscript, Weihong Deng\textsuperscript{*}\\
Beijing University of Posts and Telecommunications\\
{\tt\small \{wangmei1, whdeng, jnhu\}@bupt.edu.cn}}

\maketitle
%\thispagestyle{empty}

%%%%%%%%% ABSTRACT
\begin{abstract}
Racial equality is an important theme of international human rights law, but it has been largely obscured when the overall face recognition accuracy is pursued blindly. More facts indicate racial bias indeed degrades the fairness of recognition system and the error rates on non-Caucasians are usually much higher than Caucasians. To encourage fairness, we introduce the idea of adaptive margin to learn balanced performance for different races based on large margin losses. A reinforcement learning based race balance network (RL-RBN) is proposed. We formulate the process of finding the optimal margins for non-Caucasians as a Markov decision process and employ deep Q-learning to learn policies for an agent to select appropriate margin by approximating the Q-value function. Guided by the agent, the skewness of feature scatter between races can be reduced. Besides, we provide two ethnicity aware training datasets, called BUPT-Globalface and BUPT-Balancedface dataset, which can be utilized to study racial bias from both data and algorithm aspects. Extensive experiments on RFW database show that RL-RBN successfully mitigates racial bias and learns more balanced performance for different races.
\end{abstract}

%%%%%%%%% BODY TEXT
\section{Introduction}

Recently, with the emergence of deep convolutional neural networks (CNN) \cite{krizhevsky2012imagenet,simonyan2014very,szegedy2015going,he2016deep,hu2017squeeze}, the performance of face recognition (FR) \cite{wang2018deep,sun2014deep,schroff2015facenet} is dramatically boosted. However, as its wider and wider application, its potential for unfairness is raising alarm \cite{pmlr-v81-buolamwini18a,alvi2018turning,MITREVIEW,NYTIMES}. For instance, Amazon¡¯s Rekognition Tool incorrectly matched the photos of 28 U.S. congressmen with the faces of criminals, especially the error rate was up to 39\% for non-Caucasian people; according to \cite{garvie2016perpetual}, a year-long research investigation across 100 police departments revealed that African-American individuals are more likely to be stopped by law enforcement. As stated in the Universal Declaration Human Rights \cite{assembly1948universal}, all are equal before the law and are entitled without any discrimination to equal protection. Obviously, it is particularly important to address racial bias in existing face recognition systems. The development and deployment of fair and unbiased FR systems is crucial to prevent any unintended side effects and to ensure the long-term acceptance of these algorithms. %The recognition accuracy is not only aspect to attend when designing learning algorithms, it is further worthwhile to ensure that FR systems are fair.
Previous studies \cite{wang2019racial,zou2018ai} have shown that this bias comes on both data and algorithm aspects. Unfortunately, there are still no sufficient research efforts on the fairness of face recognition algorithms \cite{phillips2003face,phillips2011other,furl2002face}, as well as building balanced databases \cite{wang2019racial}, in the literature.

\begin{figure}[htbp]
\centering
\includegraphics[width=8cm]{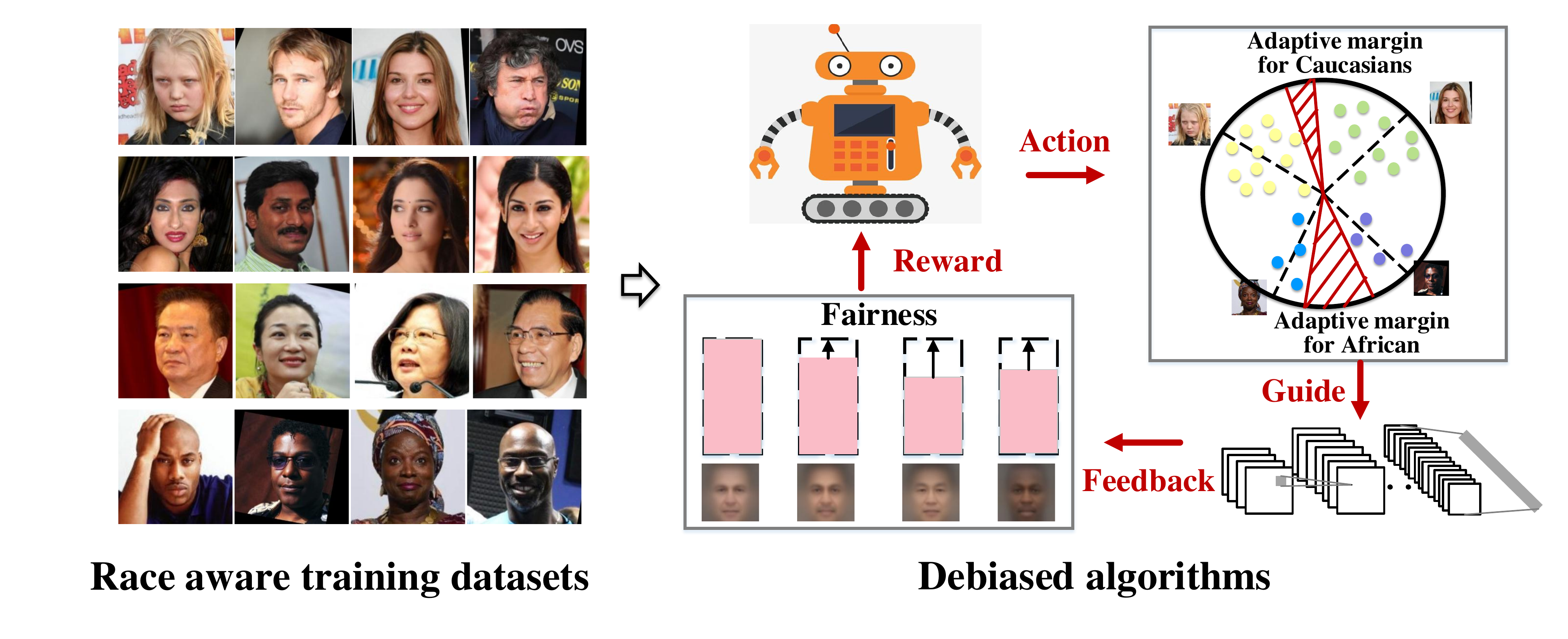}
\caption{ We provide two ethnicity aware training datasets and a debiased algorithm to reduce bias from data and algorithm aspects. }
\label{fig1}
\end{figure}

A major driver of bias in face recognition, as well as other AI tasks, is the training data. Deep face recognition networks are often trained on large-scale training datasets, such as CASIA-WebFace \cite{yi2014learning}, VGGFace2 \cite{cao2017vggface2} and MS-Celeb-1M \cite{guo2016ms}, which are typically constructed by scraping websites like Google Images. Such data collecting methods can unintentionally produce data that encode gender, ethnic and culture biases. Thus, social awareness must be brought to the building of datasets for training. In this work, we take steps to ensure such datasets are diverse and do not under represent particular groups by constructing two new training datasets, i.e. BUPT-Globalface and BUPT-Balancedface dataset. One is built up according to the population ratio of ethnicity in the world, and the other strictly balances the number of samples in ethnicity.

Another source of bias can be traced to the algorithms themselves. The state-of-the-art (SOTA) face recognition methods, such as Sphereface \cite{liu2017sphereface}, Cosface \cite{wang2018cosface} and Arcface \cite{deng2018arcface}, apply a margin between classes to maximize overall prediction accuracy for the training data. If a specific group of faces appears more frequently than others in the training data, the method will optimize for those individuals because this boosts the accuracy on the test datasets with the same bias, e.g. LFW \cite{huang2008labeled,learned2016labeled} and IJB-C \cite{maze2018iarpa}. Further, our experiments show that even with balanced training set, the feature separability of non-Caucasians is inferior to that of Caucasians. To address this problem, the algorithms must trade-off the specific requirements of margins of various groups of people, and produce more equitable recognition performance.

In this paper, we propose a reinforcement learning based race-balance network (RL-RBN) by introducing the Markov decision process (MDP) \cite{bellman1957markovian} to adaptively find optimal margins for different races with the deep reinforcement learning method \cite{mnih2015human}. First, we use deep Q-learning to train an agent to generate adaptive margin policy for non-Caucasians through maximizing the expected sum of rewards. The rewards are designed according to the skewness of intra/inter-class distances between races. Then, we train the balanced models guided by this adaptive margin policy. Finally, RL-RBN balances the inter-class and intra-class distance among different races, and thus achieves balanced generalization ability. %Although experiments are conducted to migrate racial bias,
Besides racial bias showed in our experiments, our method can also apply to remove other demographic bias, e.g. gender and age.

Our contributions can be summarized into three aspects. 1) Two ethnicity aware training datasets are constructed and released \footnote{http://www.whdeng.cn/RFW/index.html} for the study on reducing racial bias. 2) A new debiased algorithm, RL-RBN, is proposed to learn adaptive margins to mitigate bias between different races. Reinforcement learning technique is successfully applied to learn an adaptive margin policy. 3) Extensive experiments on ethnicity aware training datasets and RFW \cite{wang2019racial} shows the effectiveness of our RL-RBN. Combining balanced training and our debiased algorithm, we obtain the fairest performance across different races.

\section{Related work}

\subsection{Racial bias in face recognition}

Some studies \cite{phillips2003face,grother2010report,furl2002face,phillips2011other,klare2012face} have uncovered that non-deep face recognition algorithms inherit racial bias from human and perform unequally on different races. The 2002 NIST Face Recognition Vendor Test (FRVT) is believed to be the first study that showed that non-deep FR algorithms suffer from racial bias \cite{phillips2003face}.
Phillips et al. \cite{phillips2011other} utilized the images of the Face Recognition Vendor Test 2006 (FRVT 2006) to conduct cross training and matching on White and Asian races, and suggested that training and testing on different races results in severe performance drop. Klare et al. \cite{klare2012face} collected mug shot face images of White, Black and Hispanic from the Pinellas County Sheriff's Office (PCSO) and concluded that the Black cohorts are more difficult to recognize for all matchers. %However, the study of racial bias still remain to be vacant in deep FR systems. So it is worthwhile to go deep into this problem and exactly know the extent to which accuracy varies for different races of faces in deep learning era.
However, few efforts were made to study racial bias in deep face recognition. Recently, Wang et al. \cite{wang2019racial} contributed a test dataset called Racial Faces in-the-Wild (RFW) database, on which they validated the racial bias of four commercial APIs and four SOTA face recognition algorithms, and presented the solution using deep unsupervised domain adaptation to alleviate this bias. But there is still vacancy in training datasets which can be used to study racial bias.

%However, few studies focus on racial bias in deep era because so little training and testing information is available. Existing racial bias databases are no longer suitable for training and evaluating deep FR algorithms due to its small scale and lack of diversity of pose, age and expression. On the other hand, commonly-used training and testing databases for deep FR, e.g. LFW \cite{huang2007labeled}, IJB-A \cite{klare2015pushing}, CASIA-WebFace \cite{yi2014learning} and VGGFace2 \cite{cao2017vggface2} don't include significant racial diversity and white subjects are overwhelmingly dominant in numbers. For example, 69.9\% of the photos in LFW dataset are Caucasian people as shown in Table \ref{tab1}. In this paper, we propose two racial databases, i.e. RFW and BUPT-Worldface, that provide images of different races in the wild to enable researchers to go deep into racial bias in deep FR systems.

\subsection{Debiased algorithms}

In many computer vision applications, there are some works that seek to introduce fairness into networks and mitigate data bias. These are respectively classified as unbalanced-training \cite{sattigeri2018fairness,calmon2017optimized,more2016survey,zhou2006training}, attribute suppression \cite{alvi2018turning,mirjalili2018gender,othman2014privacy,mirjalili2018semi} and domain adaptation \cite{kan2014domain,kan2015bi,wang2019racial,sohn2018unsupervised}. By learning the underlying latent variables in an entirely unsupervised manner, Debiasing Variational Autoencoder (DB-VAE) \cite{aminiuncovering} re-weighted the importance of certain data points while training. Calmon et al. \cite{calmon2017optimized} transformed the given dataset into a fair dataset by constraining the conditional probability of network prediction to be similar for any two values of demographic information. SensitiveNets \cite{morales2019sensitivenets} proposed to introduce sensitive information into triplet loss. They minimized the sensitive information, while maintaining distances between positive and negative embeddings.

\subsection{Deep reinforcement learning}

Mimicking humans' decision making process, reinforcement learning aims to enable the agent to decide the behavior from its experiences using a Markov decision process (MDP) \cite{littman2015reinforcement}.
%With a person being generalized to an agent, the behaviors being generalized to a set of actions, a typical reinforcement learning problem can be formulated as an agent optimizes its policy of actions by maximizing the rewards it receives from an environment. Deep reinforcement learning is a combination of deep learning and reinforcement learning, which has been applied in various tasks in recent years.
Mnih et al. \cite{mnih2015human} combined reinforcement learning with CNN and bridged the divide between high-dimensional sensory inputs and actions, resulting in human-level performance in Atari Games. %In \cite{zhang2015towards}, reinforcement learning is used for controlling a robotic manipulator with visual perception only.
Recently, reinforcement learning has been successfully applied in compute vision. Rao et al. \cite{rao2017attention} used reinforcement learning to discard the misleading and confounding frames and found the focuses of attentions for video recognition. Haque et al. \cite{haque2016recurrent} utilized reinforcement learning to identify small, discriminative regions indicative of human identity in person identification. %Tang et al. \cite{tang2018deep} and Dong et al. \cite{dong2019attention} applied reinforcement learning in action recognition. Pirinen et al. \cite{pirinen2018deep} replaced the greedy candidate object regions (RoIs) selection process in object detection with a sequential attention mechanism which is trained via deep reinforcement learning. In object tracking, Dong et al. \cite{dong2018hyperparameter} proposed a novel hyperparameter optimization method that can find optimal hyperparameters for a given sequence using an action-prediction network leveraged on continuous deep Q-Learning.
In this paper, we apply deep reinforcement learning in FR to address race balance problem.

\section{Ethnicity aware training datasets}

\begin{figure}[htbp]
\centering
\includegraphics[width=8cm]{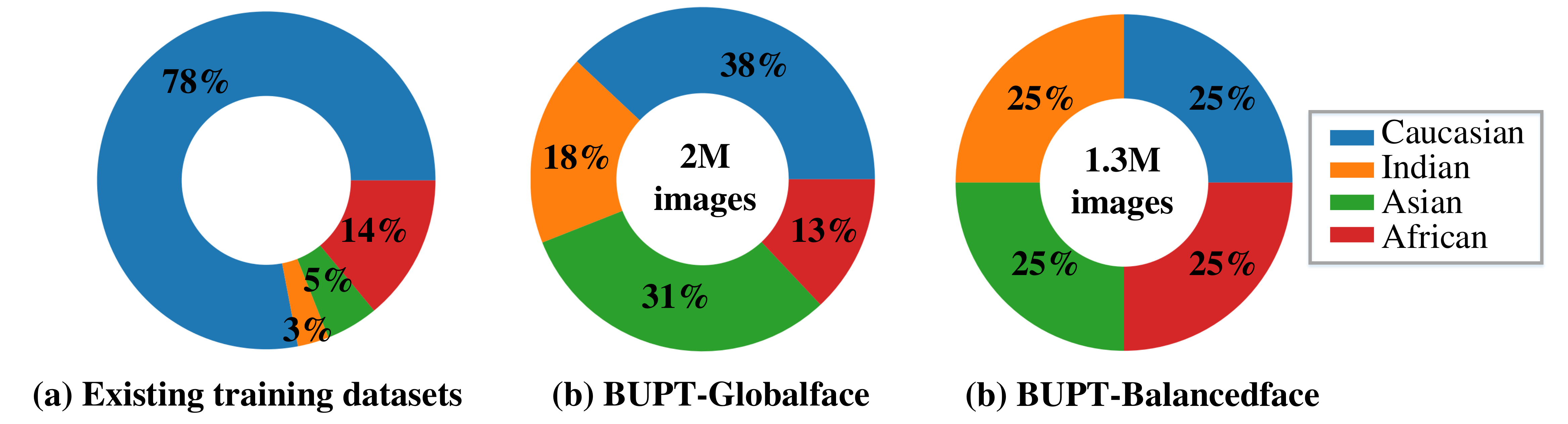}
\caption{ The percentage of different races in commonly-used training datasets, BUPT-Globalface and BUPT-Balancedface. }
\label{ratio}
\end{figure}

A major driver of bias in face recognition is the training data. Frequently, some race groups are over-represented and others are under-represented. For example, East Asia and India together contribute just 8\% of commonly-used training datasets, even though these countries represent 44\% of the world's population. %If a specific race group appears more frequently than others in the training data, the algorithms will optimize for that race because this boosts overall accuracy.
In order to remove this source of bias and represent people of different regions equally, we construct two ethnicity aware training datasets, i,e, BUPT-Globalface and BUPT-Balancedface dataset. The identities in these two datasets are grouped into 4 categories, i.e. Caucasian, Indian, Asian and African, according to their race. As shown in Fig. \ref{ratio}, BUPT-Globalface contains 2M images from 38K celebrities in total and its racial distribution is approximately the same as real distribution of world's population. BUPT-Balancedface dataset contains 1.3M images from 28K celebrities and is approximately race-balanced with 7K identities per race. %We are aware about the limitations of grouping all human ethnic origins into only 4 categories. According to studies, there are more than 5K ethnic groups in the world. We categorized the images to only 4 groups in order to maximize differences among classes.
%\begin{table}[htbp]
%	\begin{center}
%    \footnotesize
%    \setlength{\tabcolsep}{0.4mm}{
%	\begin{tabular}{c|c|c|ccccc}
%		\hline
%       \multirow{2}{*}{Database} & \multirow{2}{*}{Identities} & \multirow{2}{*}{Images} & \multicolumn{4}{c}{Racial distribution (\%)} \\
%       & & & Caucasian & Asian & Indian & African\\ \hline \hline
%       CASIA-WebFace \cite{yi2014learning} & 10K & 494K & 84.5&2.6 &1.6 &11.3  \\
%       VGGFace2 \cite{cao2017vggface2} & 9K & 3.31M &74.2 &6.0 & 4.0&15.8 \\
%       MS-Celeb-1M \cite{guo2016ms} & 100K & 10M &76.3&6.6 &2.6 &14.5 \\
%       Worldface(ours) & 38K & 2M &\textbf{38.1} & \textbf{18.3} & \textbf{30.3} & \textbf{13.3} \\ \hline
%	\end{tabular}}
%    \end{center}
%    \caption{The percentage of different race in commonly-used training databases}
%    \label{tab4}
%\end{table}

%\begin{table}[htbp]
%	\begin{center}
%    \footnotesize
%	\begin{tabular}{c|cc|cc}
%    \hline
%     \multirow{2}{*}{Subsets} & \multicolumn{2}{c|}{identity} & \multicolumn{2}{c}{image}\\
%        & number & ratio(\%) & number   & ratio(\%) \\ \hline \hline
%        %Caucasian & 14,735& 1,228,774 & 2,959 & 10,196\\
%        %Indian &7,096 &245,657 & 2,984 & 10,308\\
%        %Asian &11,749 & 406,233 & 2,492 & 9,688\\
%        %African &5,157 & 176,790 & 2,995 & 10,415 \\ \hline
%        Caucasian & 14.7K & 38.0 & 1.2M & 59.7\\
%        Indian &7.1K & 18.3 & 245.6K & 11.9\\
%        Asian &11.7K & 30.3 & 406.2K& 19.7\\
%        African &5.2K & 13.3 & 176.8K & 8.6 \\ \hline
%	\end{tabular}
%    \end{center}
%    \caption{The number of identities and images in BUPT-Worldface.}
%    \label{tab4}
%\end{table}

Similar to RFW \cite{wang2019racial}, we select images of different races from MS-Celeb-1M \cite{MSCH3} with help of the ``Nationality'' attribute of FreeBase celebrities \cite{freebase} and Face++ API. %In order to obtain the race information of each identity in MS-Celeb-1M,
%We use the ``Nationality'' attribute of FreeBase celebrities \cite{freebase} to directly select Asians and Indians. For Caucasians and Africans, Face++ API \cite{Face++} is used to estimate race. An identity will be accepted only if its most images are estimated as the same race, otherwise it will be abandoned. To avoid the negative effects caused by the biased Face++ tool, we manually check some images with low confidence scores from Face++.
However, due to unbalanced distribution of MS-Celeb-1M \cite{MSCH3}, we can only obtain 2K Indians and 5K Asians which are not enough to construct a large scale dataset. %It is not enough to construct a large scale dataset with these limited images.
%More Asian and Indian images need to be downloaded from website.
As we know, MS-Celeb-1M only selected the top 100K entities from one-million FreeBase celebrity list \cite{freebase}. %in terms of their web appearance frequency. %and released FreeBase celebrity list to encourage researchers to collect remaining data. Therefore, according to the specific names and their ``Nationality'' attribute provided by FreeBase celebrity list,
Therefore, we download the remaining images of Asians and Indians by Google according to their ``Nationality'' attribute provided by FreeBase celebrity list, and then clean them both automatically and manually. After obtaining enough images, we select images to construct our ethnicity aware training datasets, and remove their overlapping subjects in RFW \cite{wang2019racial}.

\section{Our method}

\subsection{Investigation and observation}

\begin{figure*}[htbp]
\centering
\includegraphics[width=17.5cm]{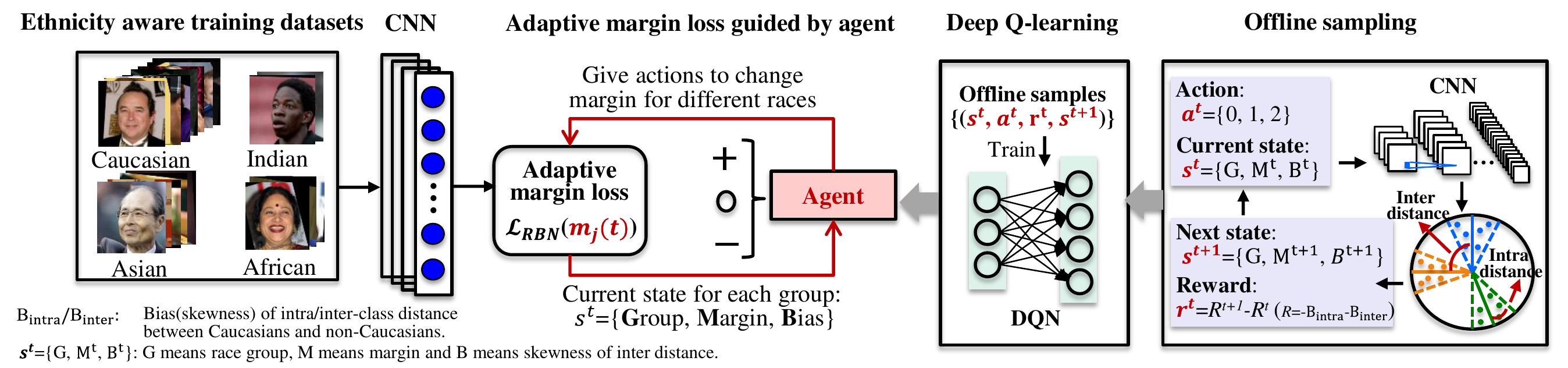}
\caption{ An illustration of our method. \textbf{Offline sampling:} We varies margin for each race group to collect some training samples , i.e. $(s^t, a^t, r^t, s^{t+1})$, before training DQN. Details of $(s^t, a^t, r^t, s^{t+1})$ refer to Section \ref{dqn}. \textbf{Deep Q-learning network:} With these samples, DQN is trained to approximate the Q-value function, and the reward is determined by the skewness of inter/intra-class distance between races. Then, adaptive margin policy for agent can be generated according to Q-value.  \textbf{Adaptive margin:} We train a race balanced network with a fixed margin for Caucasians and adaptive margins for each colored-face which changes at each training step guided by agent. }
\label{reinforcement}
\end{figure*}

%The most widely used classification loss function, softmax loss, is presented as follows:
%\begin{equation}
%L_{soft}=-\frac{1}{N}\sum_{j=1}^{N}log\left ( \frac{e^{W_{y^{\left ( j \right )}}^{T}x_{j}}}{\sum_{i=1}^{n}e^{W_{i}^{T}x_{j}}} \right )
%\end{equation}
%where $x_j\in \mathbb{R}^{d}$ denotes the deep feature of the $j$-th sample, belonging to the $y^{(j)}$-th class, and $W_i\in \mathbb{R}^{d}$ denotes the $i$-th column of the weight $W\in \mathbb{R}^{d\times n}$. $N$ is the batch size, $n$ is the number of classes. For simplicity, we fix the bias $b_i=0$ as in \cite{liu2017sphereface}. %Then, in order to optimize the feature on a sphere and make the predictions only depend on the angle between the feature and the weight, L2 normalization \cite{wang2017normface} can be applied on $W_{i}$ and $x_{j}$. Thus, the feature distances can be formulated as feature angulars: $W_{i}^{T}x_{j}=\left \| W_{i} \right \|\left \| x_{j} \right \|cos\theta _{ij}=s_c\cdot cos\theta _{ij}$, where $\theta _{ij}$ is the angle between the weight $W_{i}$ and the feature $x_{j}$ and $s_c$ is the scale factor.
To make learned features potentially separable and enhance the discrimination power, some methods proposed to incorporate a margin between classes based on Softmax. For example, Arcface \cite{deng2018arcface} used an additive angular margin $m$:

\begin{footnotesize}
\begin{equation}
L_{arc}=-\frac{1}{N}\sum_{j=1}^{N}log\frac{e^{s_c\left ( cos\left ( \theta_{y^{\left ( j \right )}j} +m \right ) \right )}}{e^{s_c\left ( cos\left ( \theta_{y^{\left ( j \right )}j} +m \right ) \right )}+\sum_{i=1,i\neq y^{\left ( j \right )}}^{n}e^{s_ccos\theta _{ij}}} \label{Arcface}
\end{equation}
\end{footnotesize}
and Cosface \cite{wang2018cosface} used an additive cosine margin $m$:

\begin{footnotesize}
\begin{equation}
L_{cos}=-\frac{1}{N}\sum_{j=1}^{N}log\frac{e^{s_c\left ( cos\left ( \theta_{y^{\left ( j \right )}j}  \right )-m \right )}}{e^{s_c\left ( cos\left ( \theta_{y^{\left ( j \right )}j}  \right )-m \right )}+\sum_{i=1,i\neq y^{\left ( j \right )}}^{n}e^{s_ccos\theta _{ij}}} \label{Cosface}
\end{equation}
\end{footnotesize}
where $\theta _{ij}$ is the angle between the weight $W_{i}$ and the feature $x_{j}$. $x_j\in \mathbb{R}^{d}$ denotes the deep feature of the $j$-th sample, belonging to the $y^{(j)}$-th class, and $W_i\in \mathbb{R}^{d}$ denotes the $i$-th column of the weight $W\in \mathbb{R}^{d\times n}$. $N$ is the batch size, $n$ is the number of classes, and $s_c$ is the scale factor.
Although large margin losses successfully improve feature discrimination, and get better performance on a series of face recognition benchmarks, Wang et al. \cite{wang2019racial} experimentally proved that they still fail to obtain balanced representations on different races. %The reasons lie in two fold: (1) Sample distribution in public datasets is uneven, where images of non-Caucasian subject are much less than Caucasian people. (2) non-Caucasians are inherently more difficult to recognize.
Insufficient samples and greater difficulty make generalization ability of non-Caucasian inferior to that of Caucasians under a uniform margin.

To get a better understanding, we train two ResNet-34 \cite{he2016deep} models with the guidance of Arcface \cite{deng2018arcface} and Softmax loss on the CAISA-Webface \cite{yi2014learning}, and give the detailed angle statistics of different races in Table \ref{angle}. The intra-class and inter-class angle are computed on set-1 and RFW \cite{wang2019racial}. The set-1 contains 500 identities per race randomly selected from our BUPT-Globalface dataset. Intra-class angle refers to the mean of angles between feature and the feature centre, which can be formulated as: $\Theta _{intra}=\frac{1}{N_g}\sum_{i=1}^{N_g}\frac{1}{\left | \mathcal{I}_i \right |}\sum_{x_{j}\in \mathcal{I}_i}\theta_{x_{j},c_{i}}$, where $N_g$ is the number of identities belonged to one race group, $\mathcal{I}_i$ is the set of all images in $i$-th identity, and $c_{i}$ is the feature centre of $i$-th identity computed by the mean vector of embeddings. Inter-class angle refers to the mean of minimum angles between embedding feature centres, which can be formulated as: $\Theta_{inter}=\frac{1}{N_g}\sum_{i=1}^{N_g}\mathop{min}\limits_{k=1:N_g,k\neq i}\theta_{c_{k},c_{i}}$. In Table \ref{angle}, we find that non-Caucasians can not obtain as good intra-class compactness and inter-class discrepancy as Caucasians do, especially the inter-class angle. %For example, with guidance of Arcface loss, the inter-class angle of Caucasians is $65.76^{\circ}$  on RFW while that of Asians is $56.46^{\circ}$.
That is to say, the performance of non-Caucasians is inferior to that of Caucasians on test set, even if a uniform margin is performed on different races when training.

\begin{table}[htbp]
	\begin{center}
    \footnotesize
	\begin{tabular}{c|c|cccc}
		\hline
         \multicolumn{2}{c|}{loss functions} & Caucasian & Indian & Asian & African  \\ \hline \hline
         \multirow{4}{*}{Softmax}&Intra1 & 35.55 & 36.00 & 38.83 & 37.29  \\
         &Inter1 & 62.67 & 50.89 & 46.17 & 54.26 \\ \cline{2-6}
         &Intra2 & 33.81 & 31.49 & 32.09 &  32.39 \\
         &Inter2 & 59.09 & 50.15 & 46.41 & 48.42  \\ \hline\hline
         \multirow{4}{*}{Arcface \cite{deng2018arcface}}&Intra1 & 36.70 & 40.00 & 42.78 & 41.50  \\
         &Inter1 & 67.72 & 59.73 & 55.34 & 62.90 \\ \cline{2-6}
         &Intra2 & 36.22 & 34.87 & 36.15 & 36.18  \\
         &Inter2 & 65.76 & 59.72 & 56.46 & 58.54  \\ \hline
	\end{tabular}
    \end{center}
    \caption{The angle statistics of different races ([CASIA \cite{yi2014learning}, ResNet34, loss*]). ``Intra1" and ``Intra2" refer to intra-class angle on set-1 and RFW, respectively. ``Inter1" and ``Inter2" refer to inter-class angle on set-1 and RFW, respectively.}
    \label{angle}
\end{table}

Considering of poorer generalization ability of non-Caucasians, we should take special care of these difficult colored faces, and thus prefer an appropriate margin with them to improve their generalization ability. Therefore, we introduce the idea of adaptive margin into race balance problem. The margin of Caucasians remains unchanged, while optimal margins are selected adaptively for each colored race in order to minimize the skewness of angles between races and learn balanced performance for different races. We replace the fixed margin $m$ in Eqn. \ref{Arcface} and Eqn. \ref{Cosface} by a race-related and training step related parameter $m_{j}(t)$, where $t$ represents the stage of the training. In fact, different races have different demands for the margins, and the demands may change during the training. The proposed adaptive margin loss function can be formulated as follows:

\begin{footnotesize}
\begin{equation}
\begin{split}
&L_{RBN}=-\frac{1}{N}\sum_{j=1}^{N}log\frac{e^{s_c\left ( cos\left ( \theta_{y^{\left ( j \right )}j} +\alpha_j(t) \right ) \right )}}{e^{s_c\left ( cos\left ( \theta_{y^{\left ( j \right )}j} +\alpha_j(t) \right ) \right )}+\sum_{i=1,i\neq y^{\left ( j \right )}}^{n}e^{s_ccos\theta _{ij}}} \label{RBN-arc}\\
&where,\ \alpha_j(t) =\left\{
             \begin{array}{lll}
             m, & if \ j\in Caucasian & \\
             m_j(t), & otherwise& \\
             \end{array}
\right.
\end{split}
\end{equation}
\end{footnotesize}
The similar modification can be made for Cosface \cite{wang2018cosface}. So the key problem is to find an optimal adaptive margin policy $m_j(t)$ for each race group to minimize the skewness of angles between Caucasians and non-Caucasians.

\subsection{Adaptive margin policy learning} \label{dqn}

In our method showed in Fig. \ref{reinforcement}, the problem of finding $m_{j}(t)$ can be formulate as a Markov Decision Process (MDP) \cite{littman2015reinforcement}, based on which we use deep Q-learning to adjust the margin at each iteration. Deep Q-learning aims to enable the agent to decide the behavior from its experiences using a MDP. %With a person being generalized to an agent, the behaviors being generalized to a set of actions, a typical reinforcement learning problem can be formulated as an agent optimizes its policy of actions by maximizing the rewards it receives from an environment.
At each time step $t\in [1,T]$, the agent takes an action $a^t$ from action space $\mathcal{A}$ according to the Q-value $Q(s^t,a)$ estimated by the deep Q-learning network (DQN) with the state $s^t$ as
input. The environment gives a reward $r(s^t,a^t)$ for this action and then the agent updates its states with $s^{t+1}$. According to this new state, the agent will go into next time step and choose a new action. %For each time step $t$, a return reward is defined as $R^t=\sum_{i=t}^{T}\gamma ^{i-t}r^i(s^i,a^i)$ where $\gamma\in (0,1]$ is a discount factor that prioritizes earlier rewards over later ones.
The goal of deep Q-learning is to train an agent with policy $\pi$ to maximize the expected sum of rewards. %Then, we give the definition of state $s^t$, action $a^t$ and reward $r^t$ in our work.

\textbf{State.} The state $s$ of the MDP consists of three separate parts \{$G,M,B_{inter}$\}. $G=\{0,1,2\}$ represents the race group, i.e. Indian (group 0), Asian (group 1) and African (group 2). $M$ is equivalent of the adaptive margin $m$. $B_{inter}$ means the bias (skewness) of inter-class distance between $g$-th race group and Caucasians, which can be formulated as follows:
\begin{equation}
\begin{split}
&B_{inter}=\left | d_{inter}^g-d_{inter}^{Cau} \right |\\
&where,\ d_{inter}^g=\frac{1}{N_g}\sum_{i=1}^{N_g}\mathop{max}\limits_{k=1:N_g,k\neq i}cos(c_{k},c_{i}) \label{diff_inter}
\end{split}
\end{equation}
where $d_{inter}^g$ and $d_{inter}^{Cau}$ is the inter-class distance of $g$-th race group and that of Caucasian, respectively. $N_g$ is the number of identities belonged to $g$-th race group and $c_{i}$ is the feature centre of $i$-th identity computed by the mean vector of embeddings. $cos(\cdot, \cdot)$ is the cosine distance function. $d_{inter}^{Cau}$ can be computed by the same way as $d_{inter}^g$. We suppose that different races have different demands for the margins, and the demands may change according to their $B_{inter}$. When the bias (skewness) is larger, the race group may need larger margin to improve its generalization ability, and vice versa. Therefore, we take both $G$ and $B_{inter}$ into consideration when designing state $s$. Each action will be chosen based on race group and the current skewness. Moreover, in order to make the space of states discrete, we map $M$ and $B_{inter}$ to discrete spaces $\mathcal{M}$ and $\mathcal{B}$, where $\mathcal{M}=\{m_1,m_2,...,m_{n_M}\}$ and $\mathcal{B}=\{b_1,b_2,...,b_{n_B}\}$.

\textbf{Action.} The action $\mathcal{A}=\{0,1,2\}$ is the adjustment of margin. We define 3 types of action as 'staying the same' (action 0), 'shifting to larger value' (action 1) and 'shifting to smaller value' (action 2), and shifting step is set to be a constant $\epsilon$. The optimal action taken by the agent at time step $t$ is formulated by $a^t=\mathop{argmax}\limits_{a}Q(s^t,a)$, where the Q-value $Q(s^t,a)$ is the accumulated rewards of taking the action $a$. For example, at time step $t$, the agent chooses to take the action '1' according to Q-value and state $s^t=\{0,m_2,d_1\}$, then the margin of Indian will update to $m_3=m_2+\epsilon$.

\textbf{Reward.} The reward, as a function $r(s^t,a^t)$, reflects how good the action taken by the agent is with regard to the state $s^t$. Since we suppose non-Caucasians should have the same generalization ability as Caucasians and the skewness between them should be minimized, we use the skewness of inter/intra-class distance between them to design the reward. The bias (skewness) of intra-class distance between Caucasians and $g$-th group can be formulated as:
\begin{equation}
\begin{split}
&B_{intra}=\left | d_{intra}^g-d_{intra}^{Cau} \right |\\
&where,\ d_{intra}^g=\frac{1}{N_g}\sum_{i=1}^{N_g}\frac{1}{\left | \mathcal{I}_i \right |}\sum_{x_{j}\in \mathcal{I}_i}cos(x_{j},c_{i})
\end{split}
\end{equation}
where $d_{intra}^g$ and $d_{intra}^{Cau}$ is the intra-class distance of $g$-th race group and that of Caucasian, respectively. $N_g$ is the number of identities belonged to $g$-th race group. $\mathcal{I}_i$ is the set of all images in $i$-th identity, and $c_{i}$ is the feature centre of $i$-th identity computed by the mean vector of embeddings. %$d_{intra}^{Cau}$ can be computed by the same way as $d_{intra}^g$.
And the bias (skewness) of inter-class distance can be computed by Eqn. \ref{diff_inter}. When the agent takes the action $a^t$ to adjust the margin of $g$-th race group, the reward of the action $a^t$ is computed by:
\begin{equation}
\begin{split}
&r(s^t,a^t)=R^{t+1}-R^t\\
&where,\ R=-\left ( B_{inter}+B_{intra} \right ) \label{reward}
\end{split}
\end{equation}

\textbf{Objective function.} We choose to use deep Q-Learning \cite{mnih2015human,watkins1992q} to learn an optimal policy for agent. A two-layer fully connected network with a further hidden layer of 10 units is utilized to estimate the Q function. Each fully connected layer is followed by a ReLU activation function. The deep Q-learning network takes the state as input and produces the Q-value of all possible actions. We update the network by the minimizing the following loss function:
\begin{equation}
\begin{split}
&L_q=\mathbb{E}_{s^t,a^t}\left [ \left ( y^t-Q\left ( s^t,a^t \right ) \right )^2 \right ]\\
&where,\ y^t=\mathbb{E}_{s^{t+1}}\left [ r^t+\gamma \mathop{max}\limits_{a^{t+1}}Q\left ( s^{t+1},a^{t+1}|s^t,a^t \right ) \right ]
\end{split}
\end{equation}
where $y^t-Q\left ( s^t,a^t \right )$ is the temporal difference error. $y^t$ is the target value of $Q\left ( s^t,a^t \right )$. $r^t$ is the reward of taking the action $a^t$, computed by Eq.\ref{reward}, and $\gamma \mathop{max}\limits_{a^{t+1}}Q\left ( s^{t+1},a^{t+1}|s^t,a^t \right )$ is the future reward estimated by the current deep Q-learning network with $s^{t+1}$.

In order to train deep Q-learning network, we collect some offline samples in advance, i.e. \{($s^t,a^t,r^t,s^{t+1}$)\}, as input to feed network. These offline samples are generated by a sample CNN which is trained by biased data. For each non-Caucasian group, we manually adjust the margin by actions $a^t=\{0,1,2\}$ and train the current sample network for one epoch with new margin. After one-epoch training, we compute the intra-class and inter-class distance of this race group and obtain the next state $s^{t+1}$ and reward $r^t$. We keep doing this until all states have been traversed. %We manually adjust the margin of different colored-faces, and train a series of sample networks with different margins in order to traverse all states and actions.
The details are presented in Algorithm \ref{al1}.

\begin{algorithm}[htb]
\caption{ Offline sampling.}
\label{al1}
\begin{algorithmic}[1]
\REQUIRE ~~\\
The unbalanced data with four race groups.
\ENSURE ~~\\
The samples, i.e. \{($s^t,a^t,r^t,s^{t+1}$)\}, which are used for training deep Q-learning network.
\FOR {$g$ \textbf{in} all groups}
\FOR {$a$ \textbf{in} all actions}
\STATE Compute $B_{inter}^t$ of $g$-th group by Eqn. \ref{diff_inter}, and obtain current state $s^t=\{G,M^t,B_{inter}^t\}$.
\STATE Take the action $a$ to adjust margin for $g$-th group.
\STATE Train the sample network for an epoch with the updated margin $M^{t+1}$ of $g$-th group; while margins of other groups remain unchanged.
\STATE Compute $B_{inter}^{t+1}$ of $g$-th group according to Eqn. \ref{diff_inter}, and obtain updated state $s^{t+1}=\{G,M^{t+1},B_{inter}^{t+1}\}$.
\STATE Compute reward $r^t$ according to Eqn. \ref{reward}.
\STATE Collect the sample \{($s^t,a^t,r^t,s^{t+1}$)\}.
\ENDFOR
\IF {$s^{t+1}$ is a new state that hasn't appeared before}
\STATE \textbf{Go} to step 2.
\ENDIF
\ENDFOR
\end{algorithmic}
\end{algorithm}

Then, we train DQN by these collected samples. After that, the adaptive margin policy can be generated according to the output of trained DQN, i.e. Q-value. We utilize the policy to guide the training process of our recognition network. The margin of Caucasians remains fixed, while optimal margins of non-Caucasians are selected adaptively according to adaptive margin policy. At each time step $t$, $s^t$ is computed by recognition network and sent to the agent, then the trained agent will take action $a^t$ to adjust the margins for each non-Caucasian group.

\section{Experiments}

\subsection{Experimental settings}

\textbf{Datasets.} %We use two datasets to train our models. One is a publicly available web-collected training dataset CASIA-WebFace \cite{yi2014learning} whose scale is small (0.5M images of 10K celebrities ). The other dataset is our BUPT-Worldface dataset, which is large-scale and contains 2M faces and 38K unique identities. Both of the two datasets have unbalanced race distribution. Our RFW dataset is utilized as test set to fairly measure performance of different race.
Because existing datasets are not race-aware except RFW \cite{wang2019racial}, we use our BUPT-Globalface and BUPT-Balancedface datasets to train our models and use RFW \cite{wang2019racial} to fairly measure performance of different races. RFW \cite{wang2019racial} consists of four testing subsets, namely Caucasian, Asian, Indian and African. Each subset contains about 10K images of 3K individuals for face verification. Moreover, in order to evaluate the generalization ability accurately, we additionally construct a validation set to compute intra-class and inter-class distance in reinforcement learning instead of directly using training sets. The validation set contains 500 identities per race, and has no overlapping subjects with BUPT-Globalface, BUPT-Balancedface and RFW dataset \cite{wang2019racial}.

\textbf{Evaluation Protocol.} Verification performance is measured by accuracy. We utilize average accuracy of four races as metric to evaluate total performance of deep models. And standard deviation (STD) and skewed error ratio (SER) are used as the fairness criterion. Standard deviation reflects the amount of dispersion of accuracies of different races. %A low standard deviation indicates that the accuracies of different races tend to be same (low bias), while a high standard deviation indicates that the accuracies of different races are spread out over a wider range (high bias).
Error skewness is computed by the ratio of the highest error rate to the lowest error rate among different races. It can be formulated as $SER=\frac{\mathop{max}\limits_{g}Error_{g}}{\mathop{min}\limits_{g}Error_{g}}$ where $g\in\{$Caucasian,Indian,Asian,African$\}$ means race group. %For example, if SE is $n$, the highest error rate is $n$ times of lowest one among different races.

\textbf{Implementation details.} For preprocessing, we use five facial landmarks for similarity transformation, then crop and resize the faces to 112$\times$112. Each pixel ([0, 255]) in RGB images is normalized by subtracting 127.5 and then being divided by 128. MxNet \cite{chen2015mxnet} is utilized to implement adaptive margin loss and TensorFlow \cite{abadi2016tensorflow} is for Deep Q-learning. We use the same CNN architecture for training recognition CNN and sampling CNN. The CNN architecture is ResNet34 \cite{he2016deep} with a modified structure, proposed in \cite{deng2018arcface}, after the last convolutional layer. CNN models are trained on two GPUs (NVIDIA GeForce 1080TI), setting the batch size as 256 for the small training dataset and 200 for the large one. For the case of training recognition CNN on the small dataset, the learning rate begins with 0.1 and is divided by 10 at the 100K, 140K iterations, and we finish the training process at 180K iterations. While training on the large dataset, we terminate at 440K iterations, with the initial learning rate 0.1 divided by 10 at the 256K, 358K, 410K iterations. We set momentum as 0.9 and weight decay as $5e-4$. The architecture of deep Q-learning is mentioned in Section \ref{dqn}. The AdamOptimizer is used to optimize the whole network. The learning rate is set to be $1e-4$ and the discount factor $\gamma$ is set to be 0.99. %During the exploration stage of the deep Q-learning, the exploration rate is decayed from 1 to 0.

First, we collect training samples \{($s^t,a^t,r^t,s^{t+1}$)\} for deep Q-learning. %We fix margin for Caucasians and varies margin for non-Caucasian groups to train sample network.
In RL-RBN(soft), the margin of Caucasians is set to be 0 similar to N-Softmax \cite{wang2017normface}, and the margin of non-Caucasians varies from 0 to 0.6 in steps of 0.2 in which an additive angular margin is used similar to Arcface \cite{deng2018arcface}; In RL-RBN(cos) and RL-RBN(arc), the margin of Caucasians is set to be 0.15 and 0.3, and the margin of non-Caucasians varies from 0.15 to 0.45 in steps of 0.1 and varies from 0.3 to 0.6 in steps of 0.1, respectively. After that, we feed the samples \{($s^t,a^t,r^t,s^{t+1}$)\} to train deep Q-learning network and then generate an adaptive margin policy for each non-Caucasian group. Finally, the recognition CNN is trained with the guidance of this policy. %in which margin for Caucasians is fixed and margin for non-Caucasian groups are adjusted by actions.

\subsection{Cause of racial bias}

\begin{table*}[htbp]
	\begin{center}
    \footnotesize
	\begin{tabular}{cc|cccc|c|cc}
		\hline
         \multicolumn{2}{c|}{Test$\rightarrow$} & \multirow{2}{*}{Caucasian} & \multirow{2}{*}{Indian} & \multirow{2}{*}{Asian} & \multirow{2}{*}{African} & \multirow{2}{*}{Avg} & \multicolumn{2}{c}{Fairness} \\
         Train$\downarrow$ & Method$\downarrow$ & & & & & & STD & SER  \\ \hline \hline
         \multirow{2}{*}{$4:2:2:2$}&N-Softmax \cite{wang2017normface} & 89.67 & 87.97 & 84.68 & 84.17 & 86.62 & \textbf{2.64} & \textbf{1.53} \\
         &RL-RBN(soft)& 91.35 & 90.77 & 89.87 & 90.13 & 90.53 & \textbf{0.66} & \textbf{1.17} \\ \hline
         \multirow{2}{*}{$5:\frac{5}{3}:\frac{5}{3}:\frac{5}{3}$}&N-Softmax \cite{wang2017normface} & 89.88 & 88.52 & 85.13 & 83.42 & 86.74 & \textbf{2.98} & \textbf{1.64} \\
         &RL-RBN(soft)&  90.33 & 90.23 & 88.97 & 89.37 & 89.73 & \textbf{0.67} & \textbf{1.22} \\ \hline
         \multirow{2}{*}{$6:\frac{4}{3}:\frac{4}{3}:\frac{4}{3}$}&N-Softmax \cite{wang2017normface} & 90.43 & 88.32 & 84.75 & 83.32 & 86.70 & \textbf{3.26} & \textbf{1.74} \\
         &RL-RBN(soft)& 90.17 & 90.02 & 87.67 & 88.27 & 89.03 & \textbf{1.25} & \textbf{1.25} \\ \hline
         \multirow{2}{*}{$7:1:1:1$}&N-Softmax \cite{wang2017normface} & 90.67 & 87.77 & 84.37 & 82.97 & 86.44 & \textbf{3.46} & \textbf{1.83} \\
         &RL-RBN(soft)& 90.63 & 90.73 & 87.72 & 87.53 & 89.15 & \textbf{1.77} & \textbf{1.35} \\ \hline
	\end{tabular}
    \end{center}
    \caption{Verification accuracy (\%) on RFW \cite{wang2019racial} trained with varying racial distribution. We boldface STD (lower is better) and skewed error ratio (SER) (1 is the best) since this is the important fairness criterion.}
    \label{simulate}
\end{table*}

\begin{figure}
\centering
\subfigure[Gaussian blur]{
\label{gau_blur} %% label for first subfigure
\includegraphics[width=3cm]{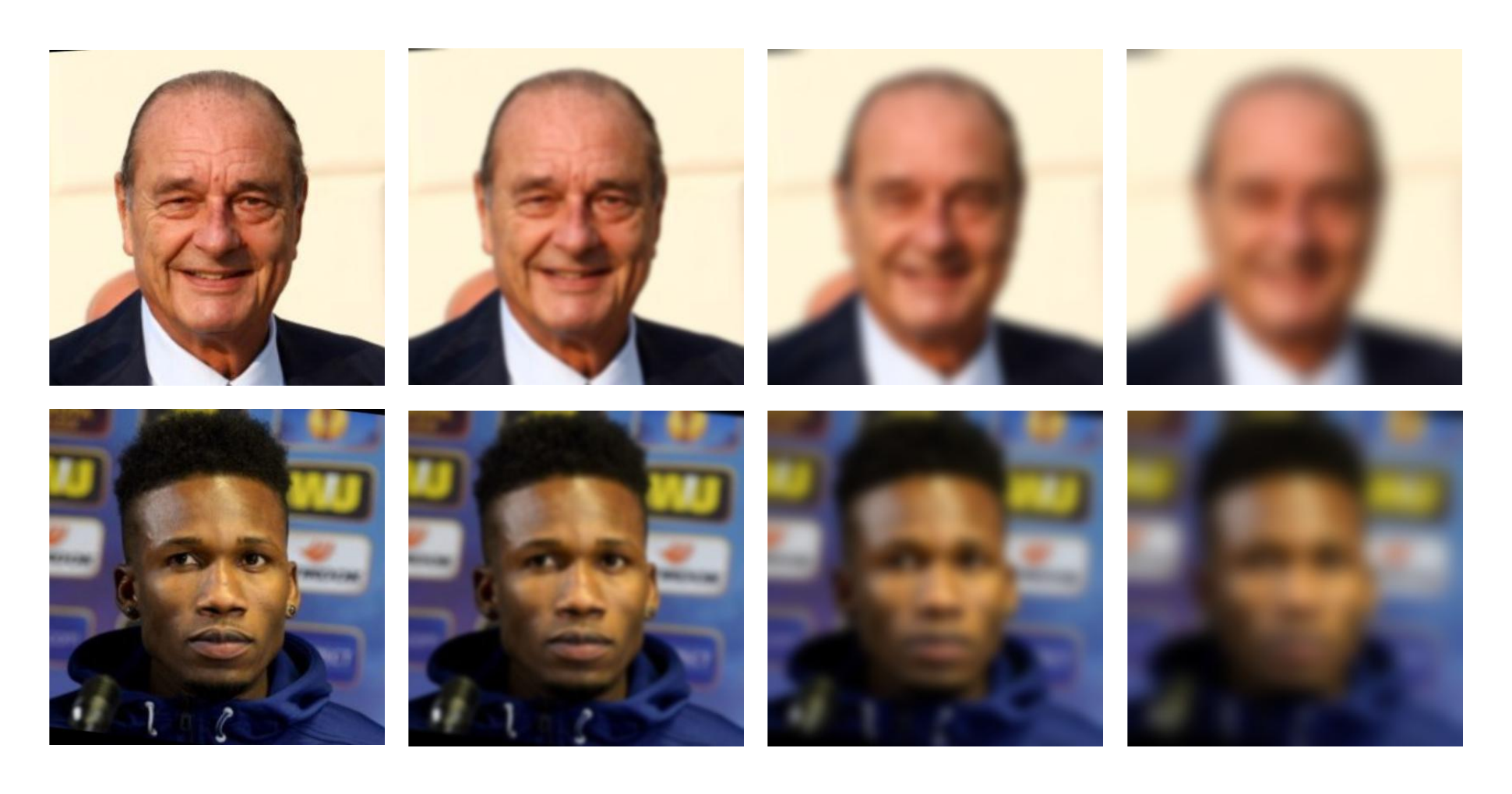}}
\hspace{0.3cm}
\subfigure[Gaussian noise]{
\label{gau_noise} %% label for second subfigure
\includegraphics[width=3cm]{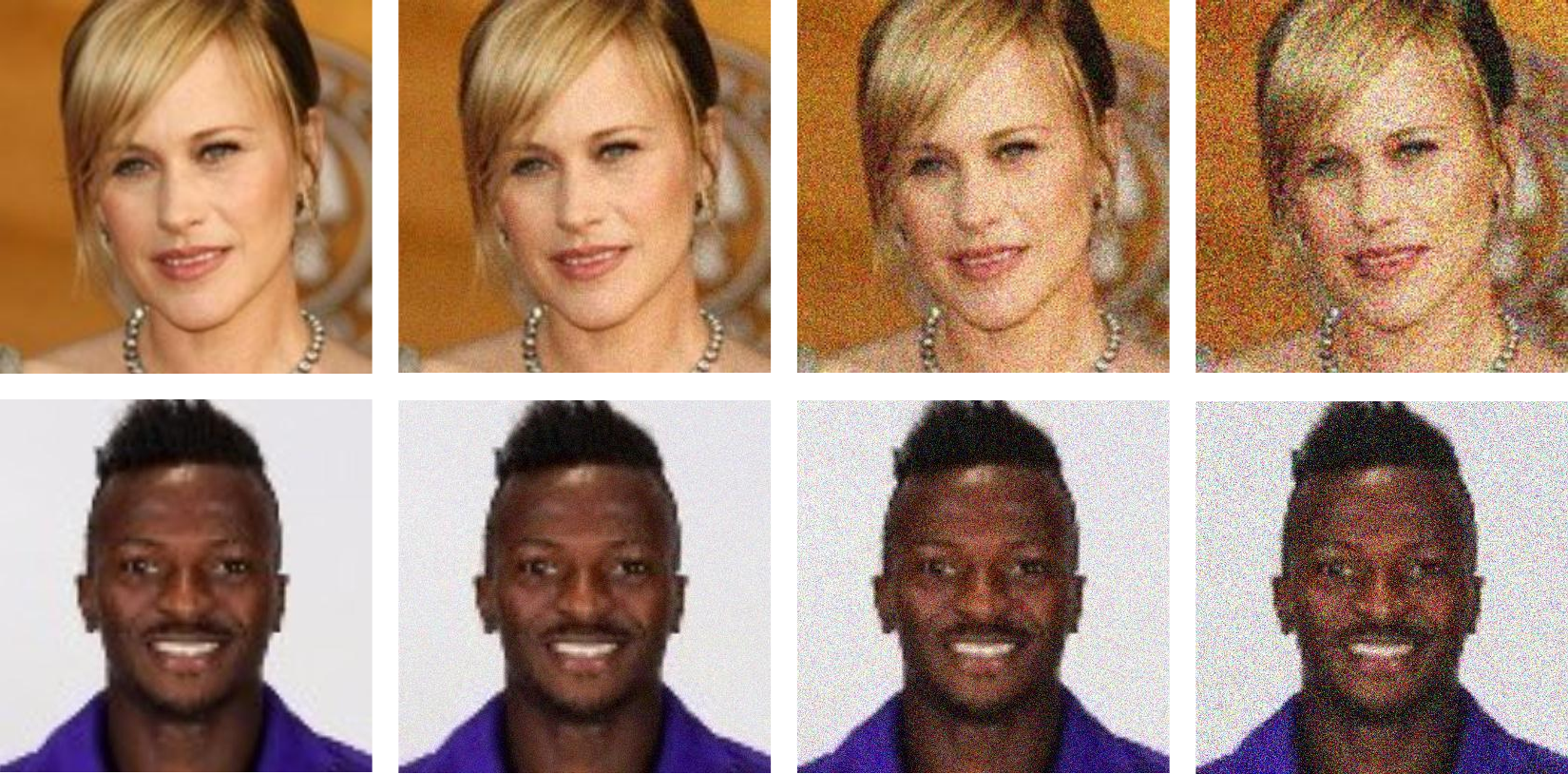}}
\caption{ Examples of images degraded by applying (a) gaussian blur or (b) gaussian noise. The first columns are original images.}
\label{gau_blur_noise} %% label for entire figure
\end{figure}

\begin{figure}
\centering
\subfigure[Gaussian blur (Softmax)]{
\label{noise-a} %% label for first subfigure
\includegraphics[width=3.5cm]{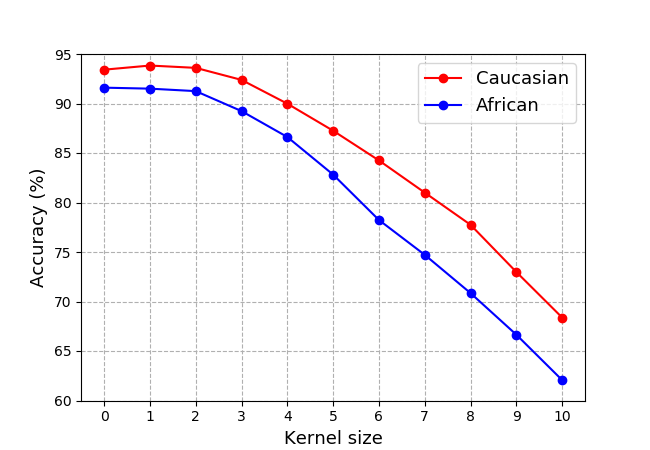}}
\hspace{0cm}
\subfigure[Gaussian noise (Softmax)]{
\label{noise-b} %% label for second subfigure
\includegraphics[width=3.5cm]{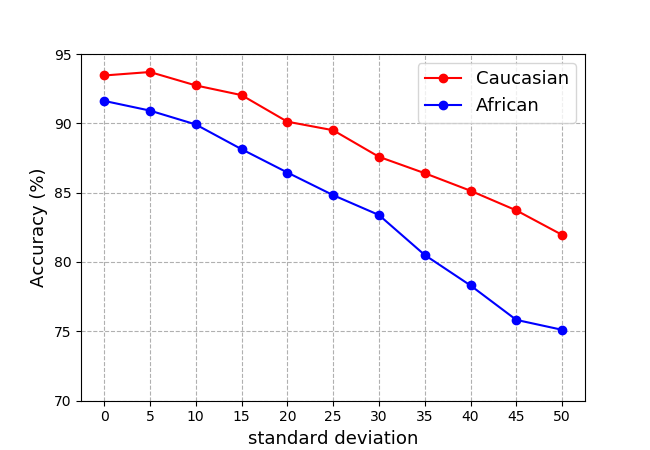}}
\hspace{0cm}
\subfigure[Gaussian blur (Arcface)]{
\label{noise-c} %% label for second subfigure
\includegraphics[width=3.5cm]{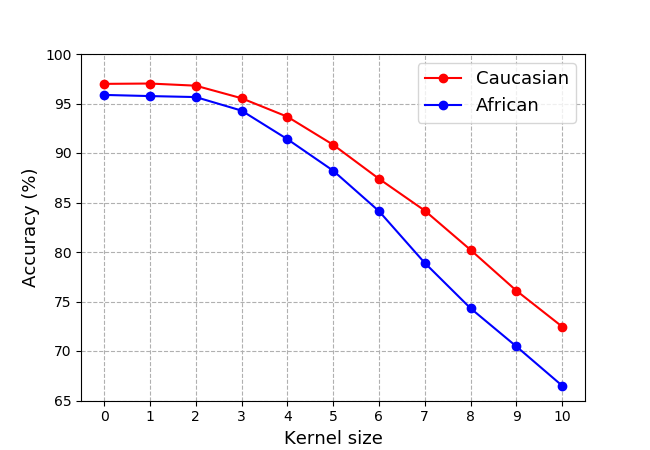}}
\hspace{0cm}
\subfigure[Gaussian noise (Arcface)]{
\label{noise-d} %% label for second subfigure
\includegraphics[width=3.5cm]{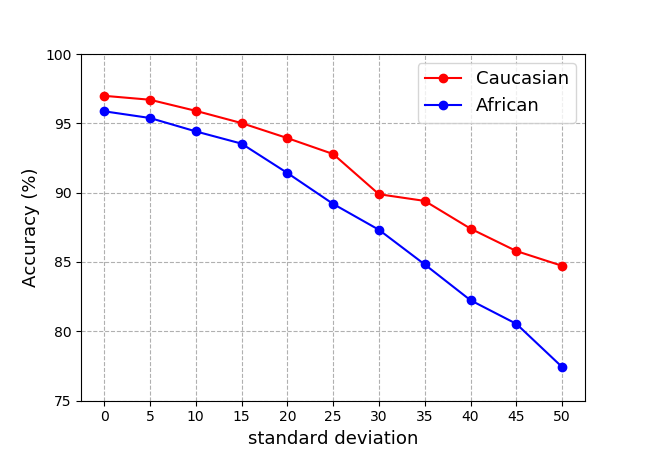}}
\caption{ The performance of Caucasians and Africans tested on blurred and noisy RFW \cite{wang2019racial}.} %The first row shows the results of model trained with Softmax loss, and the second row shows the ones trained with Arcface loss. The first column shows the results tested on blurred images and the seconde column shows the ones tested on noisy images.}
\label{finoise} %% label for entire figure
\end{figure}

Some papers \cite{wang2019racial,zou2018ai,klare2012face} verified that non-Caucasians still perform poorly than Caucasians even with balanced training and faces of colored skin are inherently difficult to recognize for existing algorithms. In order to go deep into this phenomenon, we degrade images of RFW \cite{wang2019racial}, by blur and noise, and observe the influence of these image degradations on the performance of Caucasians and Africans. To apply Gaussian blur, the sigma value of the Gaussian filter is 1.5 and kernel size varied from 1 to 10 in steps of one (Fig. \ref{gau_blur_noise}). And we add noise to images by using Gaussian noise with zero mean and varying standard deviation from 5 to 50 in steps of 5. %The examples of noisy images.
We train two ResNet-50 models with guidance of Softmax and Arcface loss \cite{deng2018arcface} using BUPT-Balancedface dataset, and test them on the blurred and noisy RFW \cite{wang2019racial}. The accuracies of Caucasians and Africans are given in Fig. \ref{finoise}. We can see that the performance gap between Africans and Caucasians still exists even with balanced training. And both Caucasians and Africans are found to be sensitive to image blur and noise. %As can be observed in Fig. \ref{finoise}, increasing blur and noise level decreases the accuracy dramatically.
More importantly, when blur and noise level increases, the performance gap between Caucasians and Africans widens. Therefore, we conclude that colored faces are more susceptible to noise and image quality than Caucasians. This may be one of the reasons why non-Caucasians are more difficult to recognize.

\subsection{Experiments of our method}

\begin{figure*}
\centering
\subfigure[Softmax]{
\label{balance_roc1} %% label for first subfigure
\includegraphics[width=2.7cm]{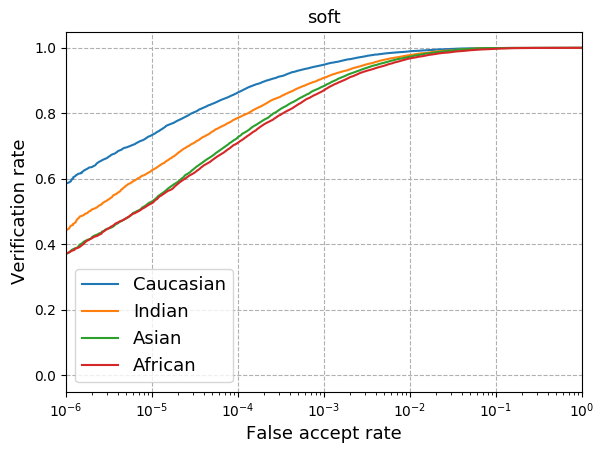}}
\subfigure[RL-RBN(soft)]{
\label{balance_roc2} %% label for second subfigure
\includegraphics[width=2.7cm]{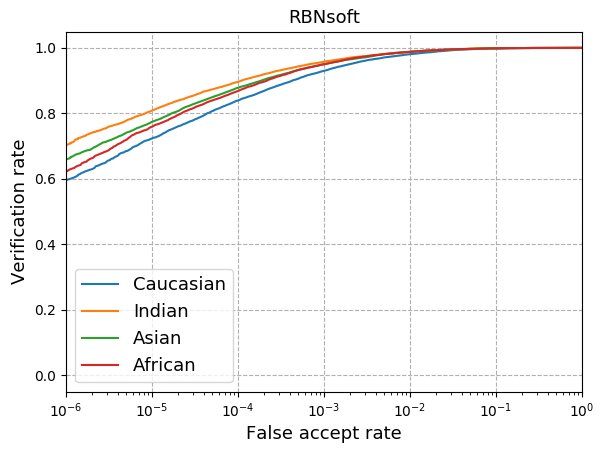}}
\subfigure[Cosface]{
\label{balance_roc3} %% label for first subfigure
\includegraphics[width=2.7cm]{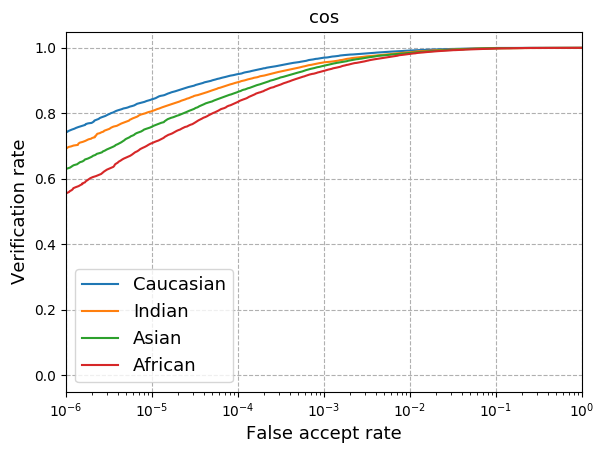}}
\subfigure[RL-RBN(cos)]{
\label{balance_roc4} %% label for second subfigure
\includegraphics[width=2.7cm]{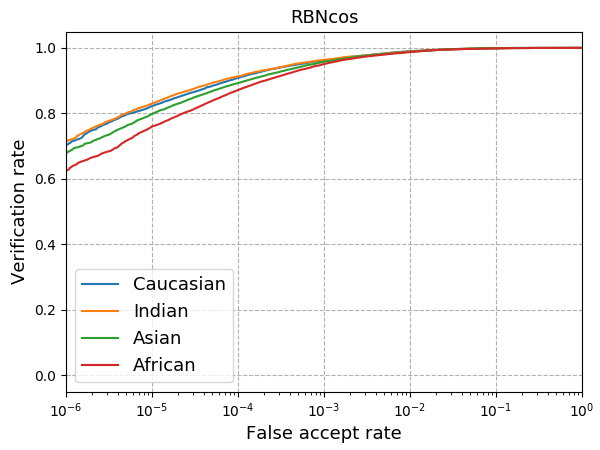}}
\subfigure[Arcface]{
\label{balance_roc5} %% label for first subfigure
\includegraphics[width=2.7cm]{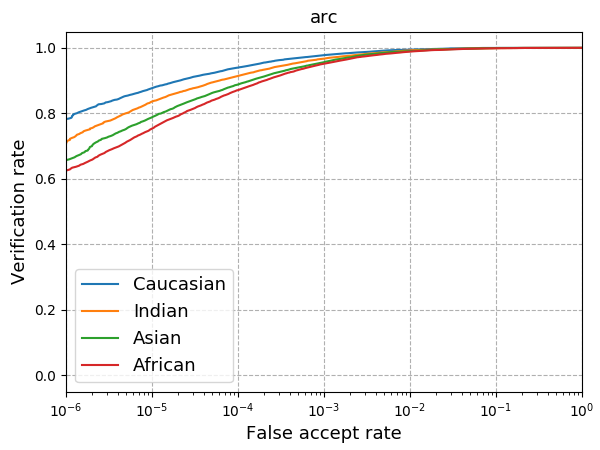}}
\subfigure[RL-RBN(arc)]{
\label{balance_roc6} %% label for second subfigure
\includegraphics[width=2.7cm]{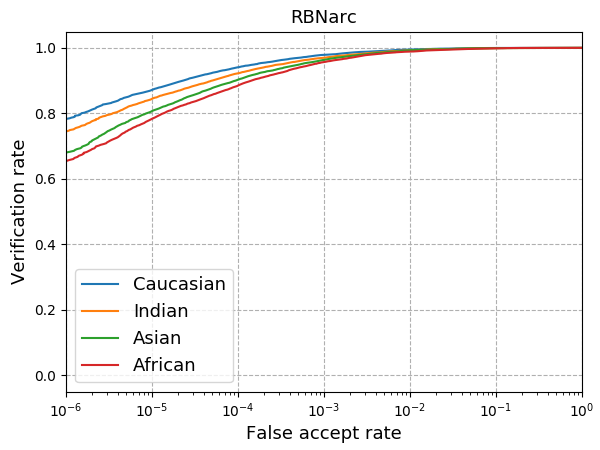}}
\caption{The ROC curves of (a) Softmax, (b) RL-RBN(soft) (c) Cosface \cite{wang2018cosface}, (d) RL-RBN(cos), (e) Arcface \cite{deng2018arcface} and (f) RL-RBN(arc) evaluated on all pairs of RFW \cite{wang2019racial}.}
\label{balance_roc} %% label for entire figure
\end{figure*}

\textbf{Results on simulated dataset.} As we know, data bias in training set severely affects fairness of algorithms. In order to validate the effectiveness of our RL-RBN, we train our algorithms using training set with different racial distribution and evaluate them on RFW \cite{wang2019racial}. We randomly pick images from our BUPT-Globalface dataset to construct these training sets. Each training set contains 12K celebrities which has the similar scale with CASIA-Webface database \cite{yi2014learning}, and is non-overlapping with RFW dataset. We make the number of Indians, Asians and Africans the same for simplicity, and change the ratio between Caucasians and non-Caucasians, i.e. \{4:6\}, \{5:5\}, \{6:4\}, \{7:3\}. %In this way, we form the training set of 4.8K2.4K2.4K2.4K, 6K2K2K2K, 7.2K1.6K1.6K1.6K and 8.4K1.2K1.2K1.2K, of which the first term (4.8K or 6K or 7.2K or 8.4K) are Caucasians and the last three terms are Indians, Asians and Africans, respectively.
Norm-Softmax \cite{wang2017normface} which normalizes weight and feature based on Softmax and is compared with our RL-RBN, as shown in Table \ref{simulate}.

From the results, we can see several important observations. First, it also shows that racial bias indeed exists in existing algorithms. For example, when the racial distribution is 4:2:2:2, the accuracy of Norm-Softmax reaches 89.69\% on Caucasian testing subset, but its accuracy dramatically decreases to 84.17\% on African subset. Second, the results quantitatively verify our thought that the accuracy of each race is positively correlated with its number in training set. For example, increasing ratio of Caucasians (from 2/5 to 7/10) increases its accuracy from 89.67\% to 90.67\% in Norm-Softmax. %By merely changing the distribution of the training set (from 4:2:2:2 to 7:1:1:1), we improve the recognition rate of Norm-Softmax by nearly 1\% on the Caucasian cohort and decrease the performance by 1.2\% on the African cohort.
Also, with the change of distribution, we observe a decrease in fairness between races. %i.e. the standard deviation changes from 2.64 to 3.46, indicative of more uneven distribution with greater racial bias.
Third, after adopting adaptive margin loss guided with reinforcement learning, our RL-RBN(soft) significantly obtains more balanced performance than Norm-Softmax on different races. %When racial distribution of training set is 4:2:2:2, the standard deviation of RL-RBN(soft) decreases from 2.64 to 0.66 compared with Norm-Softmax;
When racial distribution becomes more uneven, i.e. 7:1:1:1, our method can still perform better and decrease the SER from 1.83 to 1.35. %Benefiting from larger margin of non-Caucasians, the performances of non-Caucasians improve greatly leading to more balanced performance and better average accuracy in our RL-RBN(soft).

\begin{table}[htbp]
	\begin{center}
    \footnotesize
    \setlength{\tabcolsep}{1.1mm}{
	\begin{tabular}{c|cccc|c|cc}
		\hline
         \multirow{2}{*}{Methods} & \multirow{2}{*}{Caucasian} & \multirow{2}{*}{Indian} & \multirow{2}{*}{Asian} & \multirow{2}{*}{African} & \multirow{2}{*}{Avg} & \multicolumn{2}{c}{Fairness} \\
         & & & & & & STD & SER \\ \hline \hline
         Softmax & 95.62 & 91.97 & 90.85 & 89.98 & 92.10 & \textbf{2.48} & \textbf{2.29} \\
         M-RBN(soft) & 93.50 & 94.50 & 90.06 & 93.43 & 92.83 & \textbf{1.90} & \textbf{1.78} \\
         RL-RBN(soft)& 94.53 & 95.03 & 94.20 & 94.05 & 94.45 & \textbf{0.44} & \textbf{1.20} \\ \hline
         Cosface \cite{wang2018cosface} & 96.63 & 94.68 & 93.50 & 92.17 & 94.25 & \textbf{1.90} & \textbf{2.33} \\
         M-RBN(cos) & 96.15 & 95.73 & 93.43 & 94.76 & 95.02 & \textbf{1.21} & \textbf{1.70} \\
         RL-RBN(cos)& 96.03 & 95.15 & 94.58 & 94.27 & 95.01 & \textbf{0.77} & \textbf{1.45} \\ \hline
         Arcface \cite{deng2018arcface} & 97.37 & 95.68 & 94.55 & 93.87 & 95.37 & \textbf{1.53} & \textbf{2.33} \\
         M-RBN(arc) & 97.03 & 95.58 & 94.40 & 95.18 & 95.55 & \textbf{1.10} & \textbf{1.89} \\
         RL-RBN(arc)& 97.08 & 95.63 & 95.57 & 94.87 & 95.79 & \textbf{0.93} & \textbf{1.76} \\ \hline
	\end{tabular}}
    \end{center}
    \caption{Verification accuracy (\%) of our policy on RFW \cite{wang2019racial} ([BUPT-Globalface, ResNet34, loss*]). RL-RBN(soft), RL-RBN(cos) and RL-RBN(arc) represent our methods using adaptive margin policy based on Softmax, Cosface \cite{wang2018cosface} and Arcface \cite{deng2018arcface}, respectively. M-RBN is the method using different fixed margins for different races inversely proportional to their number.} %We boldface STD (lower is better) and skewed error ratio (SER) (1 is the best) since this is the important fairness criterion.}
    \label{worldface}
\end{table}

\textbf{Results on BUPT-Globalface dataset.} Training on BUPT-Globalface, we compare our RL-RBN with Softmax, Cosface \cite{wang2018cosface} and Arcface \cite{deng2018arcface}. The scaling parameter is set as 60 and the margin parameters are set as 0.2 and 0.3 for Cosface \cite{wang2018cosface} and Arcface \cite{deng2018arcface}, respectively. We show the results in Table \ref{worldface} and Fig. \ref{balance_roc}. First, our RL-RBN(soft) obtains more perfect performance than Softmax. It achieves about 2.35\% gains for average accuracy, and STD decreases from 2.48 to 0.44. Second, we find that large margin loss, i.e. Cosface \cite{wang2018cosface} and Arcface \cite{deng2018arcface}, can alleviate racial bias to some extent through more separate inter-class. However, racial bias cannot be eliminated completely. Third, our RL-RBN(cos) and RL-RBN(arc) can find an optimal margin for each race group and obtain more balanced performance than Cosface and Arcface. It shows the superiority of our algorithm on learning balanced features from a biased dataset.

\begin{table}[htbp]
	\begin{center}
    \footnotesize
    \setlength{\tabcolsep}{1.1mm}{
	\begin{tabular}{c|cccc|c|cc}
		\hline
         \multirow{2}{*}{Methods} & \multirow{2}{*}{Caucasian} & \multirow{2}{*}{Indian} & \multirow{2}{*}{Asian} & \multirow{2}{*}{African} & \multirow{2}{*}{Avg} & \multicolumn{2}{c}{Fairness} \\
         & & & & & & STD & SER \\ \hline \hline
         Softmax & 94.18 & 92.82 & 91.23 & 91.42 & 92.41 & \textbf{1.38} & \textbf{1.51} \\
         RL-RBN(soft)& 94.30 & 94.13 & 93.87 & 94.45 & 94.28 & \textbf{0.20} & \textbf{1.08} \\ \hline
         Cosface \cite{wang2018cosface} & 95.12 & 93.93 & 92.98 & 92.93 & 93.74 & \textbf{1.03} & \textbf{1.45} \\
         RL-RBN(cos)& 95.47 & 95.15 & 94.52 & 95.27 & 95.10 & \textbf{0.41} & \textbf{1.21} \\ \hline
         Arcface \cite{deng2018arcface} & 96.18 & 94.67 & 93.72 & 93.98 & 94.64 & \textbf{1.11} & \textbf{1.65} \\
         RL-RBN(arc)& 96.27 & 94.68 & 94.82 & 95.00 & 95.19 & \textbf{0.93} & \textbf{1.42} \\ \hline
	\end{tabular}}
    \end{center}
    \caption{Verification accuracy (\%) of our policy on RFW \cite{wang2019racial} ([BUPT-Balancedface, ResNet34, loss*]).} %RL-RBN(soft), RL-RBN(cos) and RL-RBN(arc) represent our methods using margin adaptive strategy based on Softmax, Cosface \cite{wang2018cosface} and Arcface \cite{deng2018arcface}, respectively.} %We boldface STD (lower is better) and skewed error ratio (SER) (1 is the best) since this is the important fairness criterion.}
    \label{equalizedface}
\end{table}

\textbf{Results on BUPT-Balancedface dataset.} We also compare our RL-RBN with Softmax, Cosface \cite{wang2018cosface} and Arcface \cite{deng2018arcface} on BUPT-Balancedface. The results are shown in Table \ref{equalizedface}. With balanced training, Softmax, Cosface and Arcface indeed obtain more balanced performance compared with trained on biased data. So training equally on all races can help to reduce racial bias to some extent. This conclusion is coincident with \cite{wang2019racial,zou2018ai}. However, even with balanced training, we see that non-Caucasians still perform poorly than Caucasians because some specific races are difficult to recognize. When combining our debiased algorithm and balanced data, we can obtain the fairest performance. %The SER of RL-RBN(soft) can achieve 1.08 and the SER of RL-RBN(arc) can achieve 1.42.

\begin{figure}[htbp]
\centering
\includegraphics[width=8cm]{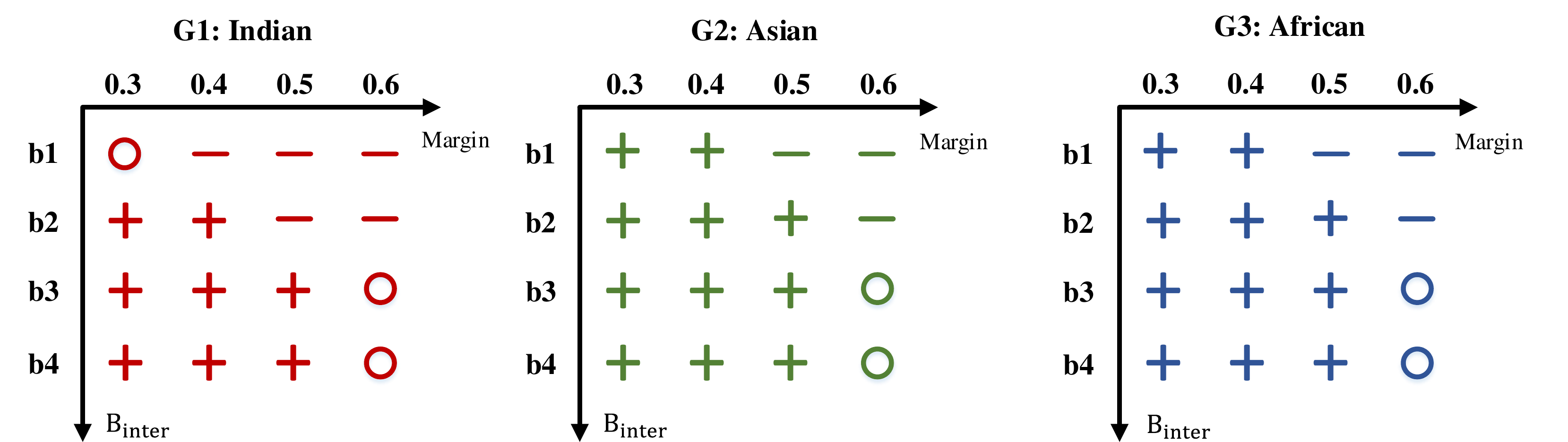}
\caption{The adaptive margin policy of RL-RBN(arc) from the trained agent. Each symbol, i.e. 'o', '+', '-', indicates an action $a=\{0,1,2\}$ based on current state $s=\{G,M,B_{inter}\}$. $M$ is mapped to discrete spaces $\{0.3, 0.4, 0.5, 0.6\}$. $B_{inter}$ is mapped to four discrete values $b1<b2<b3<b4$.}
\label{strategy}
\end{figure}

\textbf{Adaptive margin policy.} In our RL-RBN, the adaptive margin policy is given by a trained agent which can output an action $a=\{0,1,2\}$ with a state $s=\{G,M,B_{inter}\}$ as input. Here, we illustrate the adaptive margin policy of RL-RBN(arc) in Fig. \ref{strategy}. %$G$ is the race group, i.e. Indian, Asian and African. $M$ is equivalent of the adaptive margin and is mapped to discrete spaces $\{0.3, 0.4, 0.5, 0.6\}$. $D_{inter}$ means the difference of inter-class distance between each race group and Caucasians and is mapped to four discrete values $d1<d2<d3<d4$.
From the policy, we can see several important observations. First, Asian and African group have larger possibility of increasing their margin compared with Indian group. This is consistent with our theoretical analysis that we prefer stricter constraints for races which are more difficult to recognize. Based on our experiments, Asians and Africans have larger domain discrepancy with Caucasians, and perform much worse even with balanced training. %They really need a larger margin to improve their generalization ability than Indians.
Second, it is more likely for the state with larger $B_{inter}$ to increase the margin, and vice versa. Large $B_{inter}$ usually reflects less balanced performance between Caucasians and this race group so that a larger margin is supposed to improve the generalization ability of this group.

\begin{figure}
\centering
\subfigure[RL-RBN(soft)]{
\label{margin-soft} %% label for first subfigure
\includegraphics[width=3.8cm]{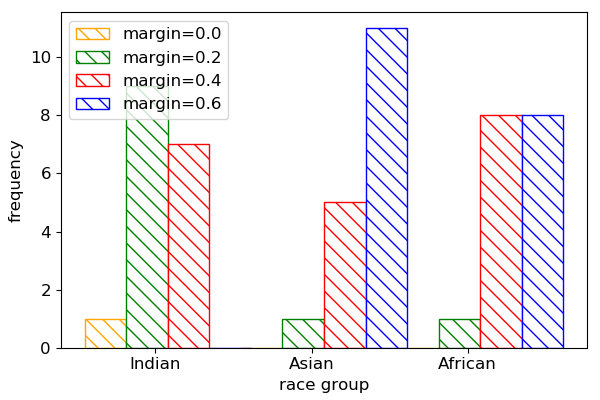}}
\hspace{0cm}
\subfigure[RL-RBN(arc)]{
\label{margin-arc} %% label for second subfigure
\includegraphics[width=3.8cm]{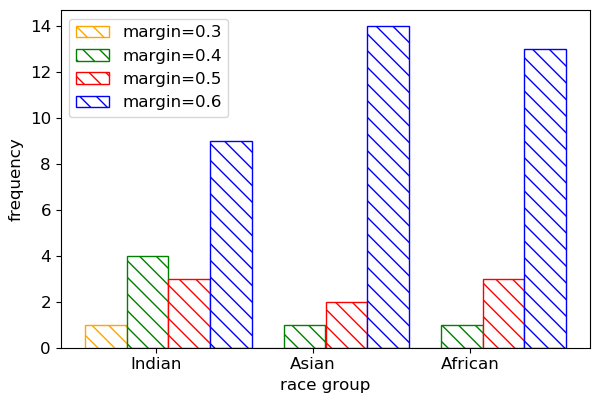}}
\caption{ Distribution of margins of non-Caucasians in RL-RBN(soft) and RL-RBN(arc) training on BUPT-Balancedface.}
\label{margin-dis} %% label for entire figure
\end{figure}

\textbf{Distribution of margins.} In Fig. \ref{margin-dis}, we illustrate the distributions of margins of non-Caucasians in RL-RBN(soft) and RL-RBN(arc) when trained on BUPT-Balancedface dataset. Guided by the agent, the margins of Asians and Africans are indeed larger than those of Indians, especially the Asians who are most difficult to recognize when trained with Softmax or Arcface. Moreover, RL-RBN(arc) usually selects larger margins for non-Caucasians compared with RL-RBN(soft). This is because that the performance of Caucasians is set to be the anchor and the performances of other races are improved to get close to them in our method. In RL-RBN(soft), the margin of Caucasians is small and fixed at 0, overlarge margins of non-Caucasians will lead to out-of-balance performance again. This result proves the robustness and adaptability of our method.

\textbf{Compared with manual margin.} We also compare our method with manual-margin based RBN (M-RBN). The M-RBN simply sets different fixed margins for different races inversely proportional to the number of their samples. From Table \ref{worldface}, we can see that the performance of Asians is always a drag on fairness in M-RBN and our method is superior to M-RBN in fairness. This is because racial bias is a complex problem in which the number is not only fact affecting out-of-balance accuracy. Although the number of Asians is much larger than that of Indians and Africans in BUPT-Globalface dataset, this group still needs a larger margin because it is the most difficult race to recognize even with balanced training.

%\begin{table}[htbp]
%	\begin{center}
%    \footnotesize
%    \setlength{\tabcolsep}{1.1mm}{
%	\begin{tabular}{c|cc|c|cc}
%		\hline
%         \multirow{2}{*}{Methods} & \multirow{2}{*}{Caucasian} & \multirow{2}{*}{non-Caucasian} & \multirow{2}{*}{Avg} & \multicolumn{2}{c}{Fairness} \\
%         & & & & STD & SER \\ \hline \hline
%         Softmax & 94.18 & 92.82 & 92.41 & \textbf{1.38} & \textbf{1.51} \\
%         RL-RBN(soft)& 94.30 & 94.13 & 94.28 & \textbf{0.20} & \textbf{1.08} \\ \hline
%         Cosface \cite{wang2018cosface} & 95.12 & 93.93 & 93.74 & \textbf{1.03} & \textbf{1.45} \\
%         RL-RBN(cos)& 95.47 & 95.15 & 95.10 & \textbf{0.41} & \textbf{1.21} \\ \hline
%         Arcface \cite{deng2018arcface} & 96.18 & 94.67 & 94.64 & \textbf{1.11} & \textbf{1.65} \\
%         RL-RBN(arc)& 96.27 & 94.68 & 95.19 & \textbf{0.93} & \textbf{1.42} \\ \hline
%	\end{tabular}}
%    \end{center}
%    \caption{Verification accuracy (\%) of our strategy trained with different loss function ([Race-Balanced dataset, ResNet34, loss*]).}
%    \label{equalizedface}
%\end{table}

\section{Conclusion}

%An ultimate face recognition algorithm should perform perfectly and fairly on different demographic group. First,
In this paper, we provide two ethnicity aware training datasets, i.e. BUPT-Globalface and BUPT-Balancedface dataset. Then, a reinforcement learning based race-balance network is proposed to alleviate racial bias and learn more balanced features. It introduces the Markov decision process to adaptively find optimal margins for non-Caucasians with the deep reinforcement learning. The comprehensive experiments prove the effectiveness of our RL-RBN.

{\small
\bibliographystyle{ieee_fullname}
\bibliography{egbib}
}

\end{document}